\documentclass[journal, twoside]{IEEEtran} 

\usepackage{multicol}
\usepackage{graphicx}
\usepackage{amssymb}
\usepackage{amsmath}
\DeclareMathOperator*{\argmax}{arg\,max}

\usepackage{calc}
\usepackage{array}
\usepackage{optidef}
\usepackage{mathtools}
\usepackage{mathrsfs}
\usepackage{bm}
\usepackage{cite}

\usepackage{booktabs,makecell,tabularx}

\newcolumntype{C}[1]{>{\centering\arraybackslash}p{#1}}
\newcolumntype{L}{>{\raggedright\arraybackslash}X}

\usepackage{siunitx}
\usepackage{etoolbox}
\newrobustcmd{\B}{\bfseries}

\usepackage{comment}
\usepackage{algorithm}
\usepackage{algpseudocode}
\usepackage{enumitem} 

\usepackage{colortbl}     
\usepackage{subfigure}
\usepackage{color, soul}
\usepackage{float}
\usepackage{multirow}
\usepackage{midfloat}
\usepackage{booktabs}
\usepackage{slashbox}
\usepackage{optidef}
\usepackage{hyperref}
\definecolor{LightBlue}{rgb}{0.75,0.936,1.00}
\sethlcolor{LightBlue}
\definecolor{LightCyan}{rgb}{0.88,1,1}

\begin{document}
\title{Learning RL-Policies for Joint Beamforming Without Exploration: A Batch Constrained Off-Policy Approach}

\author{Heasung Kim*\thanks{*Equal contribution} and Sravan Kumar Ankireddy*

\thanks{H. Kim is with the Department of Electrical and Computer Engineering, The University of Texas at Austin. Email: heasung.kim@utexas.edu}
\thanks{S. Ankireddy is with the Department of Electrical and Computer Engineering, The University of Texas at Austin. Email: sravan.ankireddy@utexas.edu}

}
\newtheorem{corollary}{Corollary}
\newtheorem{lemma}{Lemma}
\newtheorem{theorem}{Theorem}
\newtheorem{remark}{Remark}
\newtheorem{condition}{Condition}
\maketitle

\begin{abstract}
   In this work, we consider the problem of network parameter optimization for rate maximization. We frame this as a joint optimization problem of power control, beam forming, and interference cancellation. We consider the setting where multiple Base Stations (BSs) communicate with multiple user equipment (UEs). Because of the exponential computational complexity of brute force search, we instead solve this non-convex optimization problem using deep reinforcement learning (RL) techniques. Modern communication systems are notorious for their difficulty in exactly modeling their behavior. This limits us in using RL-based algorithms as interaction with the environment is needed for the agent to explore and learn efficiently.
Further, it is ill-advised to deploy the algorithm in the real world for exploration and learning because of the high cost of failure. In contrast to the previous RL-based solutions proposed, such as deep-Q network (DQN)  based control, we suggest an offline model-based approach. We specifically consider discrete batch-constrained deep Q-learning (BCQ) and show that performance similar to DQN can be achieved with only a fraction of the data without exploring. This maximizes sample efficiency and minimizes risk in deploying a new algorithm to commercial networks. We provide the entire project resource, including code and data, at the following link: \url{https://github.com/Heasung-Kim/safe-rl-deployment-for-5g}.

\end{abstract}
\begin{IEEEkeywords}
Wireless communications, coordinate multipoints, 
\end{IEEEkeywords}

\section{Introduction}
\label{sec_int}
    The goal of this project is to develop a reinforcement learning-based policy for the control problem that aims to maximize transmission rate and minimize interference in the 5G Network. The multiple base stations (BS) design wireless signals for maximizing information transmission rate over the wireless channel while the signal should not affect the signal from the other BS.
    
    For many large-scale control problems in wireless communications, reinforcement learning algorithms have been applied widely to obtain significant improvements. Combined with deep neural networks, it is possible to model and approximate many continuous state and action spaces. However, modifying the service policy of a commercial network or trying a new service policy is very risky. These RL-based algorithms cannot be directly deployed because of the initial degradation in user service, which might hurt the customer relations of the service provider. Hence, to train the RL agent without directly deploying in real-world scenarios, we need an accurate environment model. Because of the large number of components involved, especially with 5G systems becoming more prevalent, accurately modeling the environment and the movements of users is intractable. To solve this practical problem, we aim to develop a policy that learns from a finite set of data, where  interaction with the environment is severely limited. 
    
    
    In contrast, some of the previous solutions to this problem of network parameter optimization are based on the  assumptions of having access to an ideal simulator and a large number of interactions with the environment \cite{mismar2019deep}.  The authors here also assume that arbitrary radio waves (precoders) can be designed and sent for the exploration phase of a Q-learning approach. This is not feasible in practical settings; hence, we do not make this assumption in our work, which prevents our agent from exploring. We aim to develop a more practical algorithm using an offline model-based reinforcement learning that can increase sample efficiency where exploration involves a hefty cost.

\section{Background and Related Work}
In this section, we provide a brief background on wireless networks and reinforcement learning. We look at some of the relevant works that use RL-based approaches to solve wireless networking problems.

\textbf{Wireless Networks.\ }
    We consider the problem of maximizing the Signal Interference plus Noise Ratio (SINR) of a multi-user cellular communication network. We are particularly interested in  three key parameters of the setting. The first is the choice of the joint beam forming vector $\mathcal{F}$, which should increase the effective power for the intended user and decrease the interference for the other users. The second is the power $\mathcal{P}$ allotted to each user. While increasing the power of a particular user can result in improved communication links, it also increases the interference for other users and hence needs to be varied carefully. Finally, we are interested in achieving a minimum target SINR of $\gamma_{\mbox{target}}$ while maximizing the SINR and choosing the best power control and beam-forming vectors. We would like to jointly optimize all three parameters, formulated as a non-convex optimization problem, to achieve the best possible SINR.

    The authors in \cite{choi2014} discuss the problem of joint power control with the constraint of non-cooperative beam forming \textit{i.e,} the two base stations (BSs) do not exchange any channel state information (CSI) to optimize for interference cancellation but a limited number of slow-fading parameters are exchanged that will improve the SINR. The authors derived bounds on the achievable SINR and provided a scheme to improve the SINR in the presence of a matched filter (MF) based beamforming. However, this approach is general enough to be applicable to all beamforming techniques. In \cite{farr1998}, the authors propose an iterative algorithm to jointly update the power control and beam-forming weights. But a big drawback for this is the lack of consideration for the highly dynamic nature of the channel with scattering and shadowing, which are especially critical when dealing with millimeter wave settings in 5G and 6G. In the standards \cite{3gpputra}, the prevalent practice is to use an almost blank subframe (ABS) to deal with the co-channel inter-cell interference problem. But this works well only for a fixed beamforming pattern. Because of the highly dynamic nature of channels in mmWave communications, the ABS method is of limited use. Thus, as evident from the limitations of existing algorithms, we need an algorithm that can quickly adapt to the dynamic nature of the channels. 

    Reinforcement Learning is a natural choice for an online learning-based algorithm for parameter control. In \cite{luo2018}, the authors look at the beam-forming vector selection in non line of sight (NLOS) setting using a deep Q network. The sum-rate of UEs is maximized under the transmission power and minimum quality of service (QoS) constraints. A CNN-based approach was taken to estimate the Q-function. Finally, the estimated Q-function was used to employ deep Q-learning for an online control of the power allocation. Apart from power control, RL-based approaches have been used in solving other problems in the wireless domain. In  \cite{wang2018}, the authors develop a DQN-based approach to adaptively learn and find the policy that maximizes the long-term number of successful transmissions. From these works, it's evident that if used effectively, RL algorithms can provide a huge advantage over traditional algorithms in wireless communications in scenarios where analytical closed-form solutions are impossible to design. In the next section, we provide the necessary background and review of the RL algorithms that are relevant to the problem of joint power control that is being considered in this project.\\

    \textbf{Reinforcement Learning.\ }
    RL is a machine learning technique that learns the optimal set of actions to be taken to maximize the expected reward by interacting with the environment. The following are some of the key terms in RL
    \begin{enumerate}
        \item\textit{ Agent:} The learning algorithm that tries to maximize the expected reward. 
        \item \textit{State:} One of the possible scenarios that the agent can encounter during its interaction with the environment is denoted as $s \in \mathcal{S}$, where $\mathcal{S}$ is the collection of all possible states.
        \item \textit{Action:} One of the possible ways in which the agent can interact with the environment is denoted as $a \in \mathcal{A}$, where $\mathcal{A}$ is the collection of all possible actions. 
        \item \textit{Reward:} The quantified return $r$ received by the agent for taking  action $a$.
        \item \textit{Policy:} The probabilistic mapping from state $s$ to action $a$ denoted by $\pi(a|s)$.
        \item \textit{Discount factor:} The penalization term that controls the importance of future rewards, denoted by $\gamma$.
        \item \textit{Episode:} Set of all states that are part of the experience, from the starting state to the end of the interaction.
        \item \textit{Terminal State:} Special state where the episode ends or the reward is 0.
        \item \textit{Value function:} The overall expected reward starting from state $s$ assuming policy $\pi$ is followed, given by 
        \begin{align}
            V(s) = \mathbb{E}\left[\sum_{t=0}^{T-1}\gamma^{t} R_{t} | \pi \right]
        \end{align}
        \item \textit{Q-function:} The overall expected reward starting from state $s$ and action $a$, assuming policy $\pi$ is followed for future actions, given by 
        \begin{align}
            Q(s,a) = \mathbb{E}\left[\sum_{t=0}^{T-1}\gamma^{t} R_{t} | \pi \right]
        \end{align}
    \end{enumerate}


    RL algorithms for solving MDP problems are derived from tabular-based methods. Given the state transition model, the dynamic programming method of finding a solution using the contraction operator can be adopted. However, it is rare for practical problems to be given a perfect model. 
    When a transition model is not given, model-free methods that can learn without knowledge of the model can be used to find a solution. One of the most widely used model-based tabular reinforcement learning algorithms is Q-learning \cite{sutton2018reinforcement}. Q-learning approximates the action-value function using a table, which indicates a finite discrete memory that can store action-values for all the possible state-action pairs and periodically updates the approximated action-value table using the bellman operator.
    
    However, the MDP we are dealing with in this paper considers a large state space and deals with a multidimensional action space that contains all the cases we can choose from for power control and beam design. This requires tabular-based learning agents to store a large number of state-action pairs, which is not practical in terms of memory. It is also computationally expensive to obtain a desired value on the tabular storage. In such scenarios, it is often more efficient to use a function approximator to model the action-value function, which is especially beneficial for high-dimensional optimization problems. Artificial neural networks are a popular choice for function approximators and multi-layered neural networks are widely used to approximate high-dimensional state-action spaces because they have large representational capabilities \cite{lecun2015deep}.
    
    Approximation and update of action-value functions using deep neural networks are discussed in depth in \cite{mnih2015human}. The weights constituting the neural network are trained through online updates based on the Bellman equation. The authors in this paper employ various techniques to circumvent the issues encountered in practice while using the function approximators. One of the strategies is the \emph{replay memory} method, where multiple samples generated during the learning process are not discarded and instead stored in a limited size. Some of them are selected in each training process to form a batch, and the learning agent updates the weights based on that batch. Furthermore, the authors proposed the use of two neural network architectures that approximate an identical action-value function and tried reducing the variance of the action-value that is utilized as a basis for updating the weights, by fixing one of the artificial neural networks for a certain step. This resulted in a  performance beyond the expert human level in several Atari gameplay.
    
    Function approximation has been applied to policy-based reinforcement learning as well as action-value-based reinforcement learning. Using the policy gradient theorem \cite{silver2014deterministic}, it is possible to obtain a gradient of a total return, which is the sum of the reward that an agent can obtain through the interaction with an environment, with respect to the policy of the agent to maximize the return.
    The authors of \cite{lillicrap2015continuous} adopted the actor-critic method that combines the policy gradient-based method and the action-value-based method. Furthermore, the authors combined the function approximation method using an artificial neural network with the actor-critic method. The authors design actors representing policy and critics representing action-value using two different neural networks. Since the actor-network is trained with the policy gradient technique and the critic network represents an action-value function, it is updated in the direction of maximizing the MSE error based on the Bellman equation. Policy and critic representing action-value use two different neural networks. This structure showed high performance in autonomous driving, and in particular, it was shown that reinforcement learning techniques using artificial neural networks can achieve good performance even in continuous control.
    
    We will define various parameters that determine the performance of the wireless network as action. We define state as a collection of various indicators of the wireless network and the user terminals connected to the network. From this point of view, the network can be interpreted as an environment commonly referred to in reinforcement learning problems. The characteristics of this problem have additional difficulties beyond the problems of the curse of dimension for the space that we need to consider. The interaction between the learning agent and the environment requires a large cost of failure. This is because, in the wireless network system, if a new agent tries various unverified behaviors for its exploration, it can seriously impair the quality of commercial networks. To avoid performance degradation in commercial networks due to the deployment of untrained reinforcement learning agents, we can consider developing simulators to represent the commercial wireless networks, but making such a simulator that mimics a huge network also incurs a significant cost due to numerous sensitive factors that we must consider, such as user behavior patterns, signal strength, and channel environment. Hence, we can instead use the data collected previously and employ learning algorithms that do not require additional interactions with the environment. This is the reason we chose batch reinforcement learning.

    Batch reinforcement learning is a reinforcement learning approach that trains agents using only fixed batches. In \cite{fujimoto2019off}, the authors proposed batch reinforcement learning using neural network-based function approximators and the action-value and actor-critic approaches. To overcome the limitations of learning in limited batches, they introduced a generative model that helps to generate data likely to be in an already obtained batch. They also proposed a perturbation model, whose function is to take a behavior as input and re-scale it to a value within a certain interval. Such a process helps an agent trained within a limited batch to learn an appropriate behavioral pattern between imitation learning and Q-learning. These factors combined with double-Q learning \cite{fujimoto2018addressing} have been proposed as an algorithm named BCQ.
    
    We aim to train an agent to select the optimal parameters for a wireless network based on limited data by adopting BCQ approach. It is expected that the performance of the wireless network can be improved by conducting agent learning using the data already obtained without installing an unfinished algorithm in the commercial network.

\subsection{Our contributions}
    
    \begin{itemize}
        \item We propose an algorithm that attempts a model-based rollout policy at the action decision phase based on the prior knowledge of the transition model while exploiting a batch-constrained RL scheme. We call the proposed approach BCMQ which stands for Batch Constrained Model-based rollout policy Q-Learning.
        
        \item We implemented a wireless system with \emph{gym} environment class format. This system represents an environment where multiple base stations and multiple UEs can cooperate with each other to mitigate interference. This implementation allows us to fairly compare the performance of various reinforcement learning algorithms with the performance of our proposed algorithm. Through such a performance comparison, our algorithm shows that the performance of the existing state-of-the-art approaches can be approximated even with limited data given.
        
        \item We compared the performance of our proposed algorithm with the coordinate multipoint strategy using reinforcement learning, which is being intensively studied in wireless communication. We experimentally show that the performance of existing approaches using DQN can be surpassed without exploration.
        
    \end{itemize}
    To the best of our knowledge, our approach is the first 5G network parameter optimization technique using batch batch-constrained RL approach. It is expected that our approach can be usefully applied in the commercial network environment where interaction with the environment is expensive.


\section{System Model and Problem Formulation}
\label{sec_sys}

    \subsection{System Model}
\begin{figure}
        \centering
 		\includegraphics[width=1.0\linewidth]{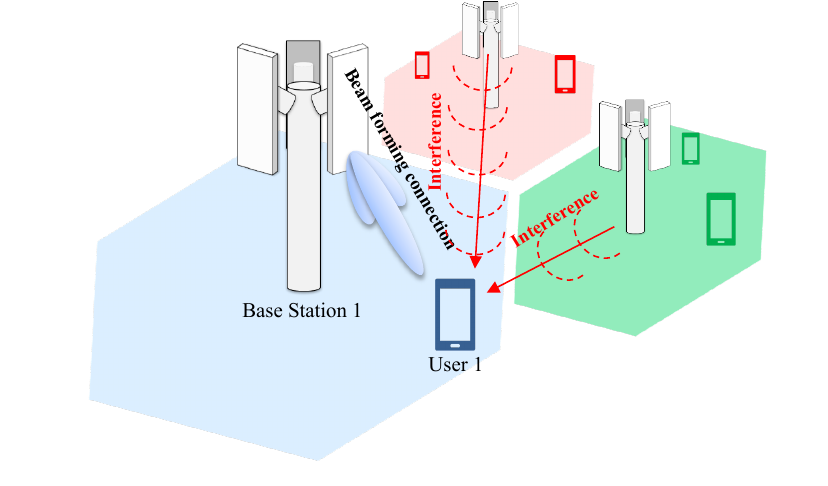}
 		\caption{ A user being served by one BS, experiences interference by other BSs. A centralized \textit{agent} performs a joint optimization over the transmit power and beam forming selection to minimize the interference and maximize the sum-rate of the users. }
     	\label{sysmodel}
    \end{figure}

        In this work, we consider the setting of multiple base stations communicating with multiple UEs, causing interference with each other, as shown in Fig. \ref{sysmodel}. We assume a centralized controller that can not only control the transmit power of each BS but also coordinate with other BSs to control the interfering transmit power as well. This is a realistic setting with the key difficulty that transmitting power that controls the quality of connection to one serving user is also the cause of degradation in quality for another user since it acts as interference. 
    
        We consider a multi-antenna set-up where each BS consists $M$ antennas and all the UEs have a single antenna. The received signal by an UE from the $l^{\mbox{th}}$ BS is given by 
        \begin{align}\label{rx_signal}
            y_l = h_{l,l}^{*} f_l x_l + \sum_{b \neq l} h_{l,b}^{*} f_b x_b + n_l,
        \end{align}
    where $x_b, x_l$ are the transmit signals corresponding to $b^{\mbox{th}}$ and $l^{\mbox{th}}$ BS respectively, while satisfying the power constraint $\mathbb{E} [|x_l|^2] = P_{TX,l}$. The $M \times 1$ beam-forming vectors, $f_b, f_l$ are chosen by the BSs in a coordinated manner to reduce the interference, based on the $M \times 1$ channels experienced by each user, $h_{l,l}, h_{l,b}$ respectively. Finally, $n_l$ is the additive white Gaussian noise that's added at the receiver according to $\mathcal{N} \sim (0,\sigma^2$. 

    \subsection{Problem formulation and terminology}
        We formulate the problem in an RL setting where the agent is not allowed to interact with the environment and can only learn from a finite set of samples \textit{i.e,} the number of samples included in the usable dataset is $N_{q} < \infty$. In other words, only $N_{q}$ queries can be accessed. We consider that this assumption is crucial for two key reasons. The first is the high cost associated with failure in commercial networks which prevents us from deploying a sub-optimal algorithm and hope to improve via exploration. The second is the fact that because of the high cost associated with data collection in real networks, we want to be as efficient as possible with the data given to us, unlike the settings where a perfect simulator is available. 
        
        We now define and revisit some of the key terms and their usage in the context of our problem formulation.
        \begin{itemize}
        
         \item \textbf{Signal-to-Interference-plus-Noise Ratio (SINR).\ } To measure the effective quality of our connection with UE, we take the ratio of the strength of our signal of interest and the sum of all unwanted signals \textit{i.e,} interference, plus noise.  This is a reliable proxy for the sum-rate capacity experienced by the user and our goal is to find the optimal parameters that can maximize this. The SINR for user $k$ in timeslot $t$ is calculated, based on Eq. \ref{rx_signal}, as 
            \begin{align}
                    \label{def_SINR}
                    \text{SINR}_{t,u} = \frac{P_{t,u} |{\mathbf{h}^{\mathrm{H}}_{t,u,b} } \mathbf{v}_{t,u} |^2}{ \sum_{j\neq k}P_{t,u} |{\mathbf{h}^{\mathrm{H}}_{t,u,b} } \mathbf{v}_{t,u} |^2 + \sigma^2_N}
                \end{align}

            where $P_{t,u=k}$ indicates a transmission power for the UE with index $k$ when time slot is $t$. The Multiple-input-single-output (MISO) channel vector between $b=k$-th BS and the UE with index $u=k$ at time slot $t$ is denoted by ${\mathbf{h}^{\mathrm{H}}_{t,u=k,b=k} }$. The noise power density is $\sigma^2_N$. Note that the signal from the undesired BS, which is not connected with the UE $k$ is considered as an interfering signal that degrades the quality of the communications. For the desired signal on the numerator of the definition (\ref{def_SINR}), ${\mathbf{h}^{\mathrm{H}}_{t,u,b} }$ has the same index $k$ for both of the UE and BS.

            \item \textbf{State.\ } We define our state as a collection of relevant information about the BS-UE connection. In this setting, we consider the transmit power and beamforming vector, represented by the index of the codebook, which is affected by the actions taken. For the setting under consideration, we consider continuous state space. We also include the geographical coordinates of the users as an observable part of the state. While the actions taken by the RL agent do not affect the coordinates of the user, these coordinates are necessary for generating the channel realization. 
            \item \textbf{Action.\ } We define our action as the change of the relevant parameters \textit{i.e.,} transmit power and beamforming a vector. For the experiments performed, we restrict ourselves to a discrete action space. \footnote{Note that our network environment is applicable to both continuous action and discrete action, but for a direct performance comparison with existing baseline algorithms, we focus on the discrete action space.} Each action taken will result in either an increase or decrease of transmit power, in multiples of pre-defined discrete steps. Similarly, the action will also result in an increase or decrease of the index for the beamforming vector. For example, consider a simple two-user case \textit{i.e,} $N_{\text{UE}} = 2$. For each user, both the increase and decrease of the transmit power and the increase and decrease of the codebook index should be determined. That is, a discrete action space of $2^2 \times 2^2 = 16$ dimension is required, which makes the problem computationally infeasible in real-time. When determining the communication method of users, the communication status of other users must be carefully considered and the status of users must be controlled at the same time so that multiple BSs can cooperate, which makes the exhaustive search even more prohibitive. 
        
            \item \textbf{Reward.\ } We define our reward as the cumulative SINR experienced by all the UEs in the system. This is reflective of our goal to maximize the sum-rate capacity of all the users. The reward is tricky in the sense that each can that results in an increased reward for a particular UE \textit{i.e.,} SINR of the UE, might result in a reduced SINR for other users. That is, this problem can be interpreted as a trade-off problem in which the powers of the desired signal and the interference signal between each UE must be well adjusted. Hence taking the cumulative sum of the rewards across all users is a reasonable representation of the system performance. 
            
            Additionally, because of extremely high variance in the channel conditions experienced by the users, the SINRs can often go to a very low or very high value, which can make learning the action-value function difficult. To circumvent this, we clip the reward to remain in between a certain range, $[SINR_{min}, SINR_{max}]$. In order to be more representative of the underlying wireless channel, average performance is measured for a large number of samples. We finally define the reward as 
            $$R_{t} = clip(\sum_{k=1}^{N_{\text{UE}}} \text{SINR}_{t,u=k}, \text{SINR}_{max},\text{SINR}_{min}).$$
            

            \item \textbf{Dynamics.\ } For the RL algorithm to be meaningful, we first need to formulate our problem in a sequential manner. In our problem, at any time step $t$ the agent can take action $a_t$ that affects the choice of transmit power and beamforming vectors in the next time step $t+1$. Because of this, the reward observed at time step $t$, $R_{t+1}$, is dependent on the action $a_t$ which in turn affects action at time $t+1$, $a_{t+1}$. Hence the transmit power and the beamforming vector vary sequentially till the end of the episode. Thus, the problem can be formulated as an MDP, and the future states are decided by the current actions taken by the RL agent. On a side note, the geographical location of the users is an observable state that is unaffected by the actions of the agent and is only essential for channel realization. 
            
            To provide more context, the RL agent can take an action to increase or decrease the transmit power for each user individually and similarly for the codebook index. The result of these actions jointly decides the transition to the next state. 
            
            
             \textbf{Our goal.\ } The goal of this problem is to maximize the discounted sum of UEs' SINR in the maximum number of slots $T$ that determines one radio frame \textit{i.e.,} we consider an episodic setting of length $T$.
            \begin{maxi}
                {\{a_{t}\}_{t=0}^{T-1}} { \mathbb{E}\left[\sum_{t=0}^{T-1}\gamma^{t}  R_{t} \right]}{}{}
                {\label{def_optimization_probelm}}{}
            \end{maxi}
            As mentioned earlier, this metric is a good proxy for the overall performance of the UEs, and optimizing this results in a high data rate in a real network. In particular, we will use and develop the batch-constrained method appropriately to solve this problem only for a given dataset without interaction with the environment. From this point of view, the expectation of the problem (\ref{def_optimization_probelm}) is calculated on the state dynamics that can be expressed from the given $N_{q}$ data samples.
        \end{itemize}

    \subsection{Complexity of the Parameter Optimization Problem}
    Without any intelligent learning-based policy, the naive way to select the parameters will be through brute force search over the parameter space. At each time instant $t$, we need to search over $\mathcal{P}$ power levels and $\mathcal{F}$ beamforming vectors, resulting in a search space of $\mathcal{P} \times \mathcal{F}$ for each user. The resultant search complexity for brute force would thus be $(\mathcal{P} \times \mathcal{F})^{N_{UE}}$. This scales rapidly with an increase in number of users, antennas, and feasible power levels, making it intractable. 
    
\begin{figure}
        \centering
 		\includegraphics[width=1.0\linewidth]{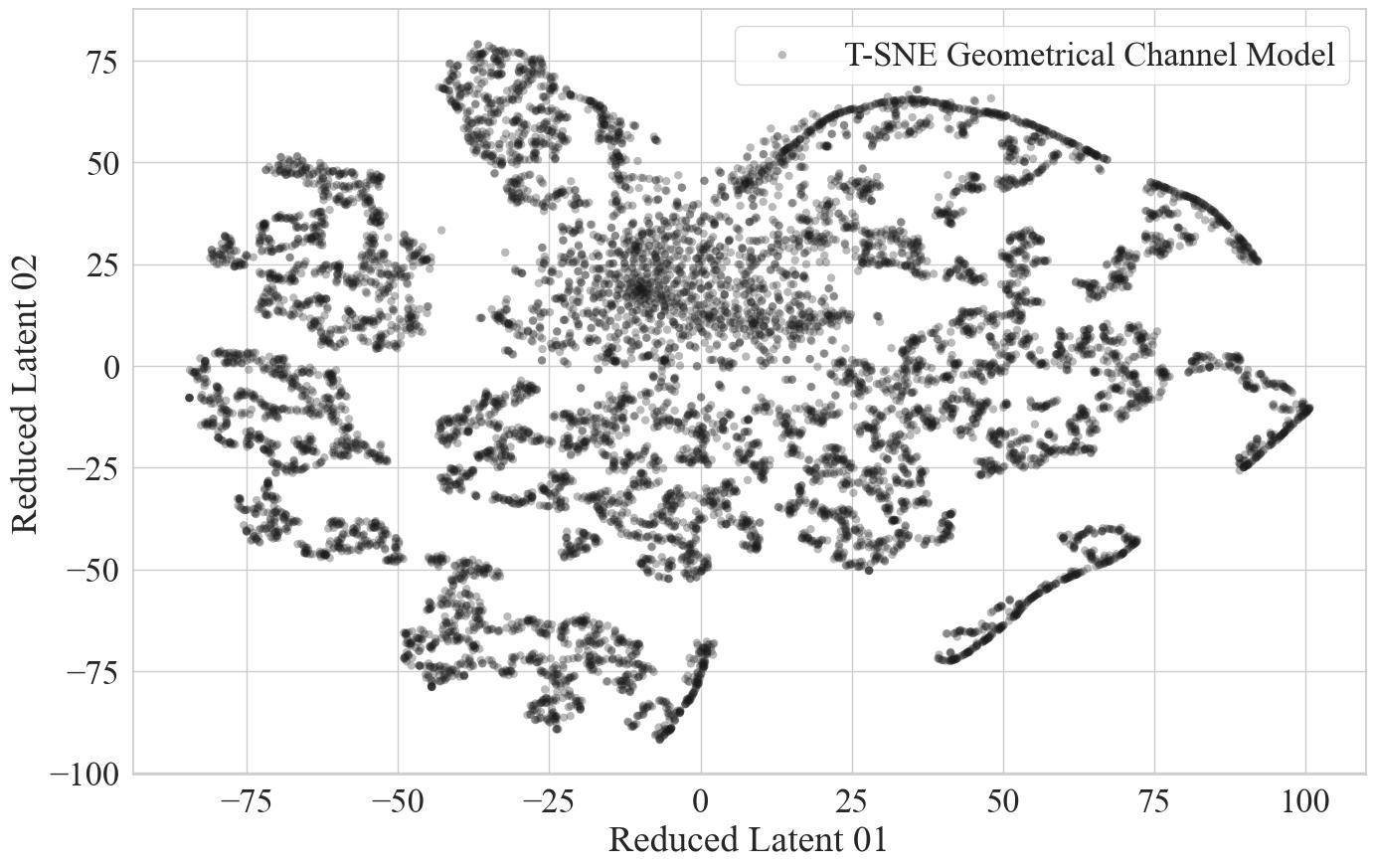}
 		\caption{A scatter plot for two 10,000 channel state realizations ${\mathbf{h}_{t,u,b} }$ on the two latent spaces from T-SNE result. The geometrical channel model follows a distribution in which the power and phase of radio waves are determined based on the location of the user and the base station. However, it is difficult to obtain a rule for selecting the optimal codeword under this distribution in closed form. }
     	\label{TSNE_channel_model}
    \end{figure}

    Even for a single UE coordinate. the variations in the resultant channel experienced are huge, making bruteforce search for nay optimal configuration impossible. Fig. \ref{TSNE_channel_model} shows a scatter plot for the two latent components from T-SNE \cite{van2008visualizing} of the channel realization from the geometrical channel distribution. Hence, this supports our argument that we need a better approach to solve this huge non-convex optimization problem and we choose to approach this as a MDP problem using reinforcement learning techniques. 

    \section{Our approach}
    
    Since our state space is continuous due to the users' coordinates and the transmission power levels, we use an approximated version of the action-value function to reduce the algorithm complexity. Moreover, even for simple 2-user settings, we saw that the action space is huge $a \in \{0,1\}^{N_{\text{UE}}\times{2 ^2} }$. Hence, using function approximation is the rational choice for this setting.
    
    As mentioned earlier, we consider a discrete action space so that we can make a direction comparison with the baseline performance of DQN from \cite{mismar2019deep}. The original BCQ algorithm introduced in \cite{fujimoto2019off} is for a continuous action space, which contributed to most of the complexity of the algorithm. For our setting here, a simpler variant of BCQ, called discrete batch-constrained deep Q-learning, introduced in \cite{fujimoto2019benchmarking} is sufficient. As the name suggests, this is a discrete action space variant of the original BCQ algorithm. Now that we decided on the algorithm of choice, we will explain our approach in more detail below.
    
    \subsection{Batch Constrained Reinforcement Learning}
    In the original BCQ, a generator $G_w$ is trained that can enable us to sample from an action that can be selected after sampling from $G_w$ and perturbing appropriately. In a discrete case, instead of this complicated procedure, we can simply compute the probabilities of every action conditioned on a state and consider the top contenders, based on a threshold as 
    \begin{align}
        \pi(s) = \mbox{argmax}_{a | G_w(a|s)/\mbox{max} \hat{a} G_w(\hat{a}|s) > \tau} Q_\theta(s,a)
    \end{align}
    
    It is important to adaptively adjust the threshold to achieve the best performance. Thus, we scale the probabilities by the maximum of all probabilities over all actions to allow only those actions whose probability is above a certain threshold. 
    
    Finally, the training of the Q function approximator network is done by formulating the policy selection as 
    \begin{align}
        \label{def_action_value_loss}
        &\mathcal{L}(\theta) \nonumber \\
        &= l_k(r +  \gamma \max_{\tiny{a' | G_w(a',s')/ \mbox{max} \hat{a} G_w(\hat{a} | s') > \tau}}  Q_\theta(s',a'), Q_\theta(s,a)),
    \end{align}
    where $l_k$ is the loss function for action-value training. We use the mean squared error loss for our training.
    Since we have 
    $G_w(s,a) \approx \pi_{\text{batch}}(s,a)$ where $\pi_{\text{batch}}$ is a policy which has been used, the trainable parameters $\Theta_{G_{w}}$ for $G_w(s,a)$ can be updated in direction of minimizing negative log-likelihood loss for the input $s,a$, and the trainable parameters $\Theta_{Q}$ of the action-value function can be updated in direction of minimizing the mean squared error. The Adam optimizer \cite{kingma2017adam} is adopted for both updates.
    \begin{figure}
        \centering
 		\includegraphics[width=1.0\linewidth]{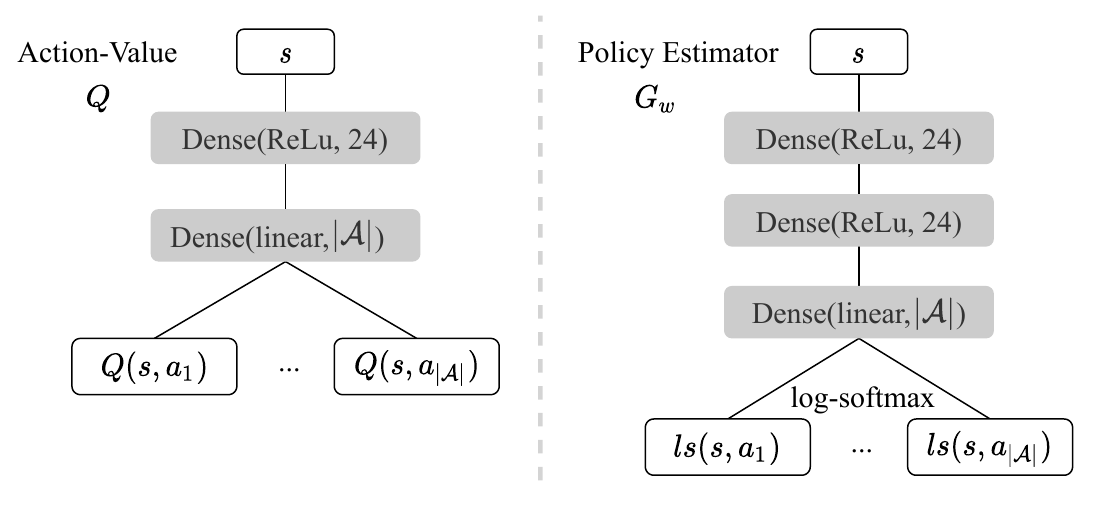}
 		\caption{Approximated action-value function $Q$ and the batch-policy estimator $G_{w}$. $ls$ indicates the log-softmax output of $|\mathcal{A}|$-sized output from the dense nerual network.}
     	\label{figure_neural_architecture}
    \end{figure}
    Fig. \ref{figure_neural_architecture} shows the overall structure of the neural network. Our state dimension can be interpreted as a latent vector of valid and significant values. To this end, we do not use a layer that aggregates information such as convolution but construct a function approximation with a dense net and activation.
    
    






    \subsection{One-step Tree Search for Action Selection}
    We started with our baseline DQN approach and replaced the algorithm with a BCQ-based learning model. Now, to further explore better algorithms, we make a simple addition to our BCQ algorithm to make in even better. We use a one-step rollout, which has been proven to make good policy even better. Since only a tiny amount of time is allowed for action decisions for the scheduling of communication systems, we use the learned $G_{w}$ and $Q$ to implement a one-step approximation rollout. This methodology does not deviate from the agent's behavior pattern learned by the off-policy policy, and at the same time, the model can be actively used. More specifically, action candidates that satisfy the condition of $ \frac{G_{w}(s,a) }{ \argmax_{a'} G_{w}(s,a')} > \tau$ are found. These are actions that exactly follow the learned agent's policy. Because our environment is partially predictable, the next states derived from these behaviors are approximately predictable. We use $k=2$, that is, the agent chooses the two most preferred actions, and then re-estimates the maximum action-value in the state derived from those actions. Through this logic, the action that can achieve the highest reward in one step future is finally selected.
    
    While we are constraining our setup to be not too complicated because of the limited time frame of the project, it is easy to understand the benefit of this addition. We call the new algorithm a Batch Constrained Q-Learning with a One-step Rollout algorithm. 
    
    We provide the algorithm in detail in Algorithm \ref{algorithm_WP}.
    
\begin{algorithm}
 \caption{Batch Constrained Q-Learning with One-step Rollout}
 \label{algorithm_WP}
 	\begin{algorithmic}[1]
    
 \renewcommand{\algorithmicrequire}{\textbf{Input:}}
 \renewcommand{\algorithmicensure}{\textbf{Output:}}
    \State Generate limited data ($N_{q}$ samples)
    \State Start Training Phase
 	\State Initialize the parameters $\Theta_{G_{w}}$. $\Theta_{Q}$ and $\Theta_{Q'}$
    \While {$i< \text{max learning iteration}$}
  	    \State Get mini-batch, $(s_{t},a_t,r_t,s_{t+1})$ pairs
        \State Adam update $\Theta_{Q}$ with MSE
        \State Adam update $\Theta_{G_{w}}$ with negative log-likelihood
        \State Soft target update  $\Theta_{Q'} = \Theta_{Q'} \tau_{s} + \Theta_{Q}(1-\tau_{s}) $
        \State $i=i+1$
  	\EndWhile
  	
  	\State Training Phase Ends.
  	
  	\State Deployment (Start Test)
  	\While {$t < T$}
  	    \State Observe state $s_{t}$
        \State Get a action set $\mathcal{A}_{\text{BCQ}}=\{a |  \frac{G_{w}(s,a) }{ \argmax_{a'} G_{w}(s,a')} > \tau \}$
        \State Top k actions $\mathcal{A}_{\text{Top-k}} = \{ a| \text{Top-k}_{a\in \mathcal{A}_{\text{BCQ}}} Q(s,a) \}$
        
        \State Expect Next states $s'$ for each actions
	    
  	    \State Action $a=\argmax_{a\in \mathcal{A}_{\text{Top-k}}} Q(s',a')$
  	    \State Apply action $a$ and get $s_{t+1}$
  	    \State $t=t+1$
  	\EndWhile
  	
  	\State End Test
  	
 	\end{algorithmic} 
\end{algorithm}

\section{Numerical Results}
    \subsection{Environment Setting}
    For performance comparison with the existing approach, we used the same network parameters to reproduce the results of \cite{mismar2019deep}, and these network settings follow the radio propagation path loss models for millimeter-wave wireless networks channel model implementation of \cite{sulyman2016directional}. The area around each BS where a UE can communicate with the BS successfully is referred to as a cell. We consider an environment with 2 cells and the two cells can cause interference to the other users.
    The $X-Y$ coordinates of the BS corresponding to the first cell is (0,0) and the BS corresponding to the second cell is located at (0,255). Throughout the work, we consider the unit of measurement as meters. The users are randomly located around the BS in a radius of 150, with the center assumed at the location of BS, at the beginning of each episode. That is, our model includes both the behavior and channel distribution of UEs communicating in an area covered by a set of specific BSs. For the radio propagation, $T_{K}=290$ Kelvins, $B=15000$Hz, and the Boltzmann constant is $1.38e-23$. $\text{SINR}_{max}$ and $\text{SINR}_{min}$ are set to $-50$ and $200$, respectively, for the clipping.
    
    
     
    \subsection{Hyperparameter Setting}
    
    The hyperparameter selection and tuning were done mostly via heuristics and turning. We began by using the same hyperparameter settings as the baseline algorithms and varied them slowly to arrive at the best set of parameters for our setting. For example, you can see from Fig. 5 that the choice learning rate affects the learning and the reward significantly. Similarly, we decided to use a batch size of 32. The actual hyperparameters used for generating each plot are included in the label and description of the figures.
    
    \subsection{Performance Evaluation}
    We repeated the same experiment 20 times with the same parameter set to clearly and accurately measure the performance of our proposed approach. This experimental approach allows us to know the overall performance of the reinforcement learning algorithm as mentioned in \cite{henderson2018deep}. This is for the sake of fair comparison among the competitors and their final performance, as well as uncertainty.
    
    \begin{itemize}
        \item \textbf{Batch Constrained Q Learning with Model-based Rollout (BCMQ, proposed)}
        BCMQ is the proposed approach we described in the previous section. Note that this algorithm only learns using the given data and no interaction with the environment is required.
        \item \textbf{Soft-Actor Critic (SAC)}
        For comparison with the State of the art algorithm, we employed the SAC of \cite{haarnoja2018soft, christodoulou2019soft}. SAC is a method of updating the policy in the direction of maximizing the value function while inducing the policy to take as many various actions as possible in entropy-based MDP. Both the value function and policy function are calculated via function approximations using a deep neural network. The training uses the standard mini-batch stochastic gradient descent (SGD) based updates in the direction of  the estimated value from the transition dynamics and the estimated value function.

        To reproduce the performance, we start with the implementation of \cite{ota2020tf2rl} and add additional modules required to incorporate changes required specific to our setting. This allows us for a fair comparison, minimizing any differences in performance caused by differences in implementation. 
        
        \item \textbf{Deep Q-Learning (DQN)}
        Deep Q-Learning \cite{mnih2015human} uses a deep neural network for the action-value function approximation. 
        For the sake of fair performance comparison, the performance of this algorithm is reproduced with the implementation in \cite{ota2020tf2rl}. \\
    \end{itemize} 
    
    \begin{figure}
        \centering
 		\includegraphics[width=1.0\linewidth]{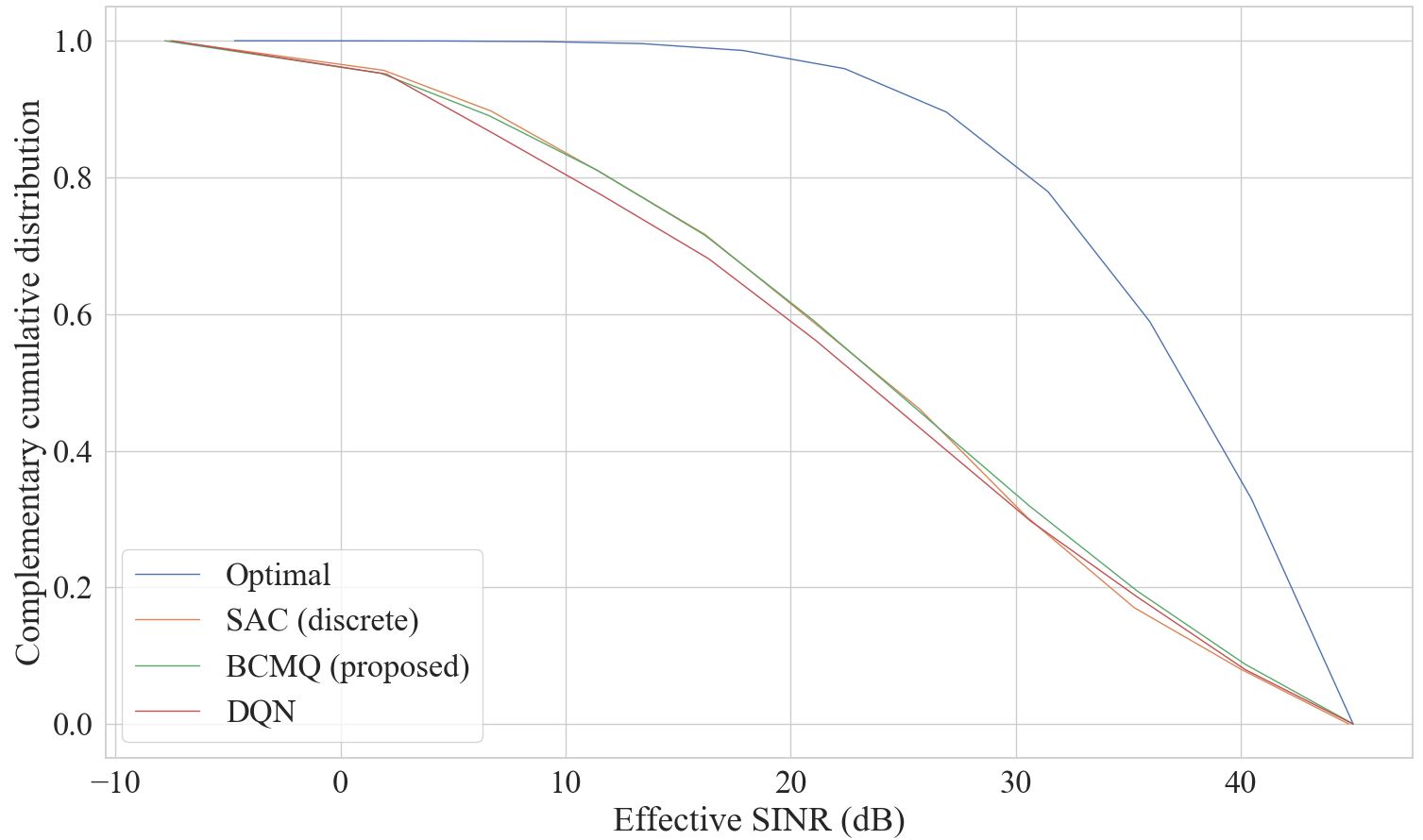}
 		\caption{Learning curves of average sum of rewards over 1,000 radio frames (episodes). The width equal to the deviation from the mean is filled with the corresponding color. In order to minimize the uncertainty due to randomness and to measure the exact performance of the algorithms, the same experiment was repeated 10 times with the same set of parameters for each of the approaches.}
     	\label{ccdf}
    \end{figure}
        
        \textbf{Absolute performance.\ } In Fig. \ref{ccdf}, we plot the Complementary Cumulative Distribution Function (CCDF) of the effective SINR experienced by the UE $\gamma_{eff}^{l}$. For any random variable, CCDF is defined as 
        \begin{align*}
            \bar{F}_X(x) = P(X > x).
        \end{align*}
        
        Thus, the CCDF of the SINR can be used as a proxy for the quality of communication experienced by users. We limit the SINR to a reasonable range of $[-5 \text{dB}, 60 \text{dB}]$. The reason for limiting the range of SINRs considered is because of the transmit power limitations and the Quality of Service (QoS) guarantees that the service provider needs to satisfy, which varies significantly because of the high variation in the channel and interference experienced by the users. Hence we only consider the simulation where the connection to a UE is valid \textit{i.e,} the SINR experienced satisfies these conditions. We discard the remaining simulations at the end of the episodes. 
        
        The plot labeled as \textit{Optimal} in Fig. \ref{ccdf} represents the maximum theoretically achievable and the goal of any policy would be to get as close to this as possible. Note that this is very hard to achieve in a practical wireless network where the channel is unknown and the complexity of brute force search is very high. We can see that in comparison with the existing baseline DQN \cite{mismar2019deep}, we see that the algorithm proposed here BCMQ is considerably better. But the most important result here is that we achieve this performance without any exploration, using only a fraction of the data of DQN. 
        
        
        
    \begin{figure}
        \centering
 		\includegraphics[width=1.0\linewidth]{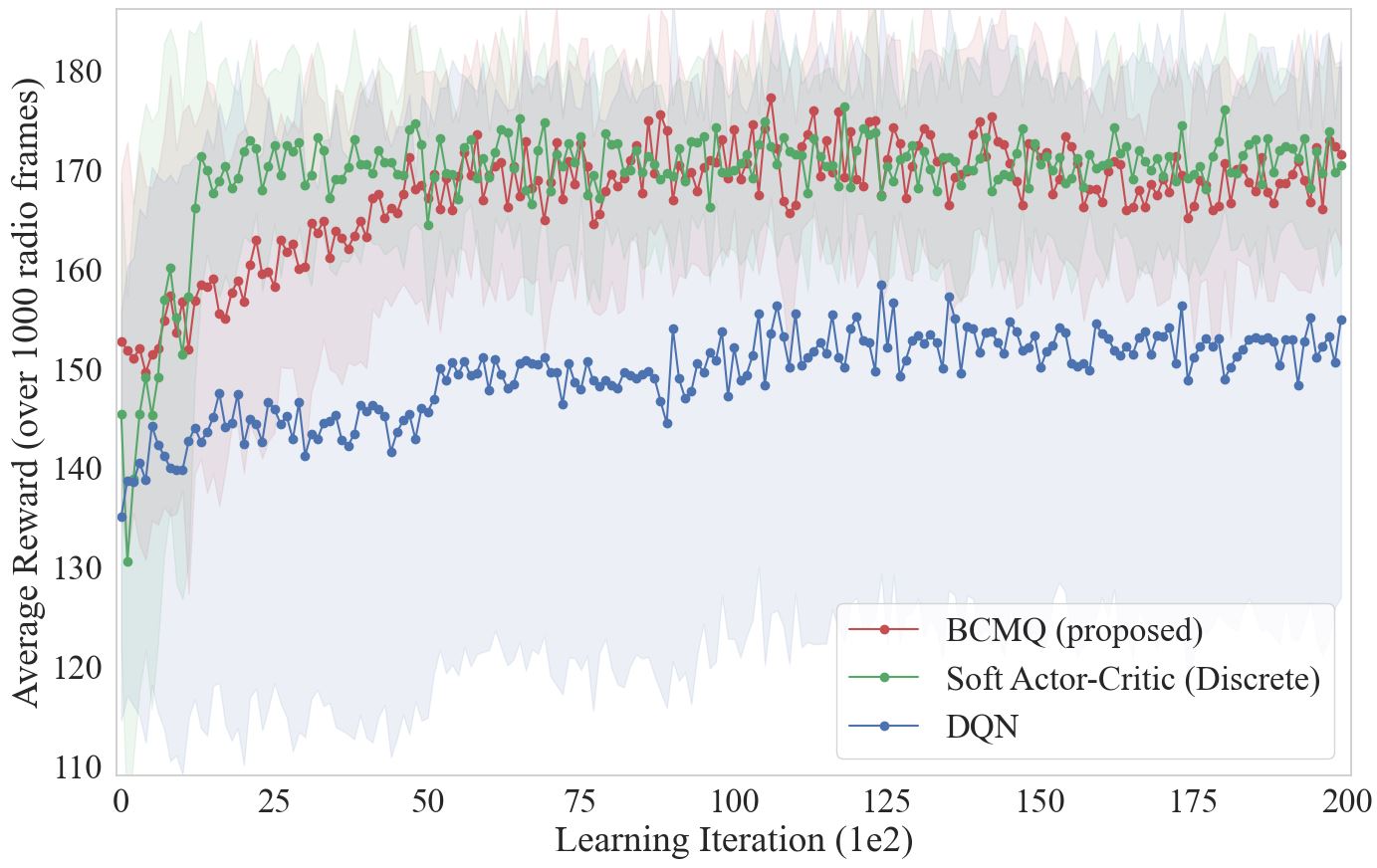}
 		\caption{Learning curves of average sum of rewards over 1,000 radio frames (episodes). The width equal to the deviation from the mean is filled with the corresponding color. In order to minimize the uncertainty due to randomness and to measure the exact performance of the algorithms, the same experiment was repeated 10 times with the same set of parameters for each of the approaches.}
     	\label{average_reward_vs_learning_iter}
    \end{figure}

        \textbf{Convergence.\ } Fig. \ref{average_reward_vs_learning_iter} shows the convergence of the algorithms BCMQ, SAC, and DQN. We chose SAC to compare the performance with the state-of-the-art reinforcement learning algorithm as a target for comparison. Additionally, to compare against the existing baseline, we chose the DQN approach proposed in \cite{mismar2019deep} as a target for comparison. It is important to note that both SAC and DQN require exploration. For our experiments, we construct a batch by choosing actions at uniform, random, over all states and generate a data set of 20,000 samples. Even with this constrained batch data set, the proposed BCQM algorithm eventually achieves performance on par with SAC. We note that despite using the same learning rate, the convergence is different for different algorithms considered, which is expected because of differences in architectures. As seen from the figure, while the convergence for SAC is faster, BCQM is also comparable in this aspect. In our opinion, for practical network settings, the pros of not having to use exploration for BCQM far outweigh the cons of slower convergence compared to SAC. 
        
    
    \begin{figure}
        \centering
 		\includegraphics[width=1.0\linewidth]{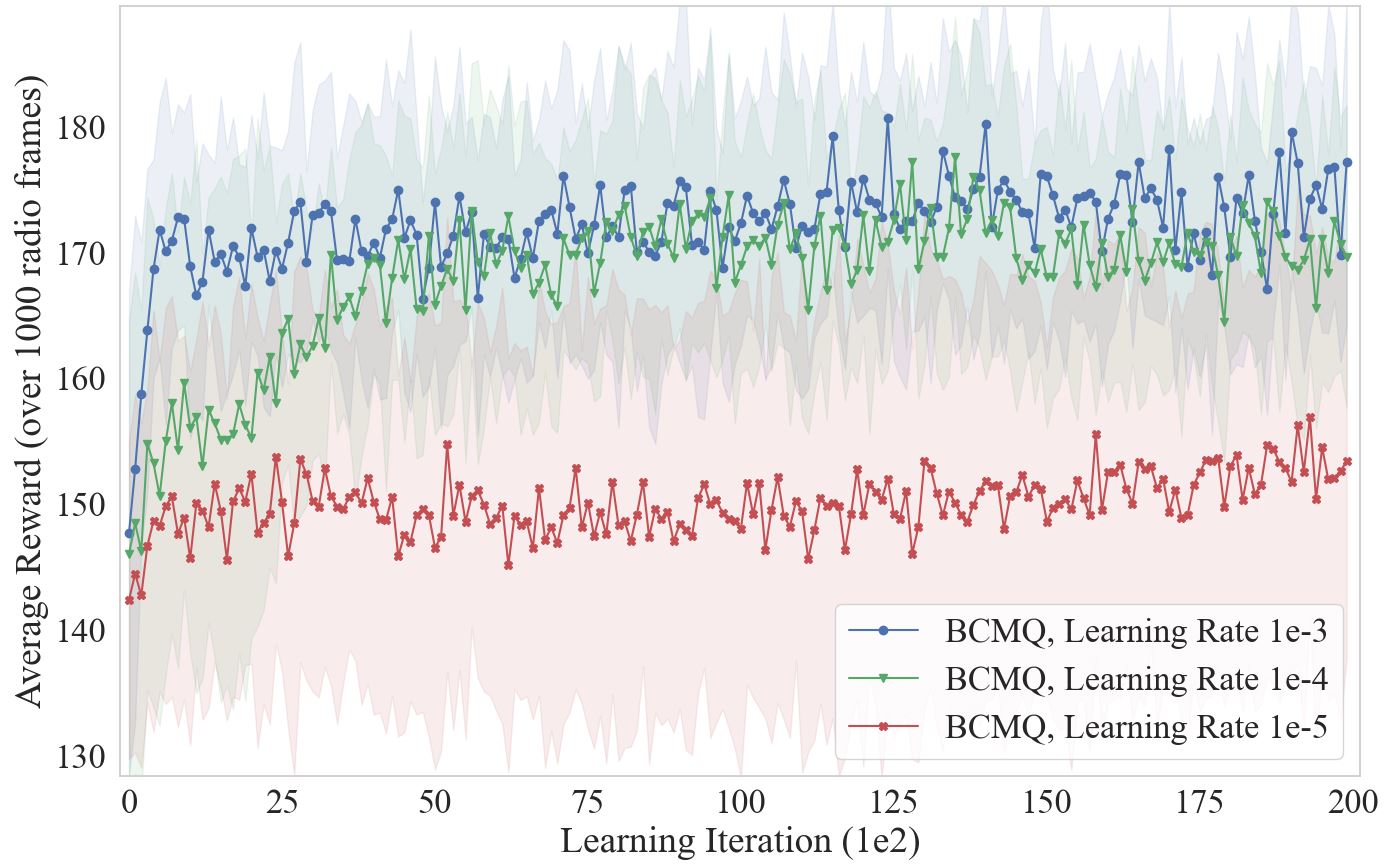}
 		\caption{Comparison of performance according to learning rate.}
     	\label{average_reward_vs_learning_iter_LRdiffer}
    \end{figure}
    
        \textbf{Learning rate.\ } In Fig. \ref{average_reward_vs_learning_iter_LRdiffer}, we plot the performance of our algorithm for some of the learning rates we tried. By varying the learning rate gradually over a range of values, we pick the best-suited rate for our model. For the purpose of representation, we plot the trends for learning rates $\{10^{-3},10^{-4}, 10^{-5} \}$. We can see from the plot that when a learning rate of $10^{-5}$ was applied, the convergence was significantly slow. Moreover, there is no significant difference in performance between a learning rate of $10^{-3}$ and $10^{-4}$, though the former is slightly more unstable. Hence we fix our learning rate as $10^{-4}$ for all of our experiments. 
        
    \begin{figure}
        \centering
 		\includegraphics[width=1.0\linewidth]{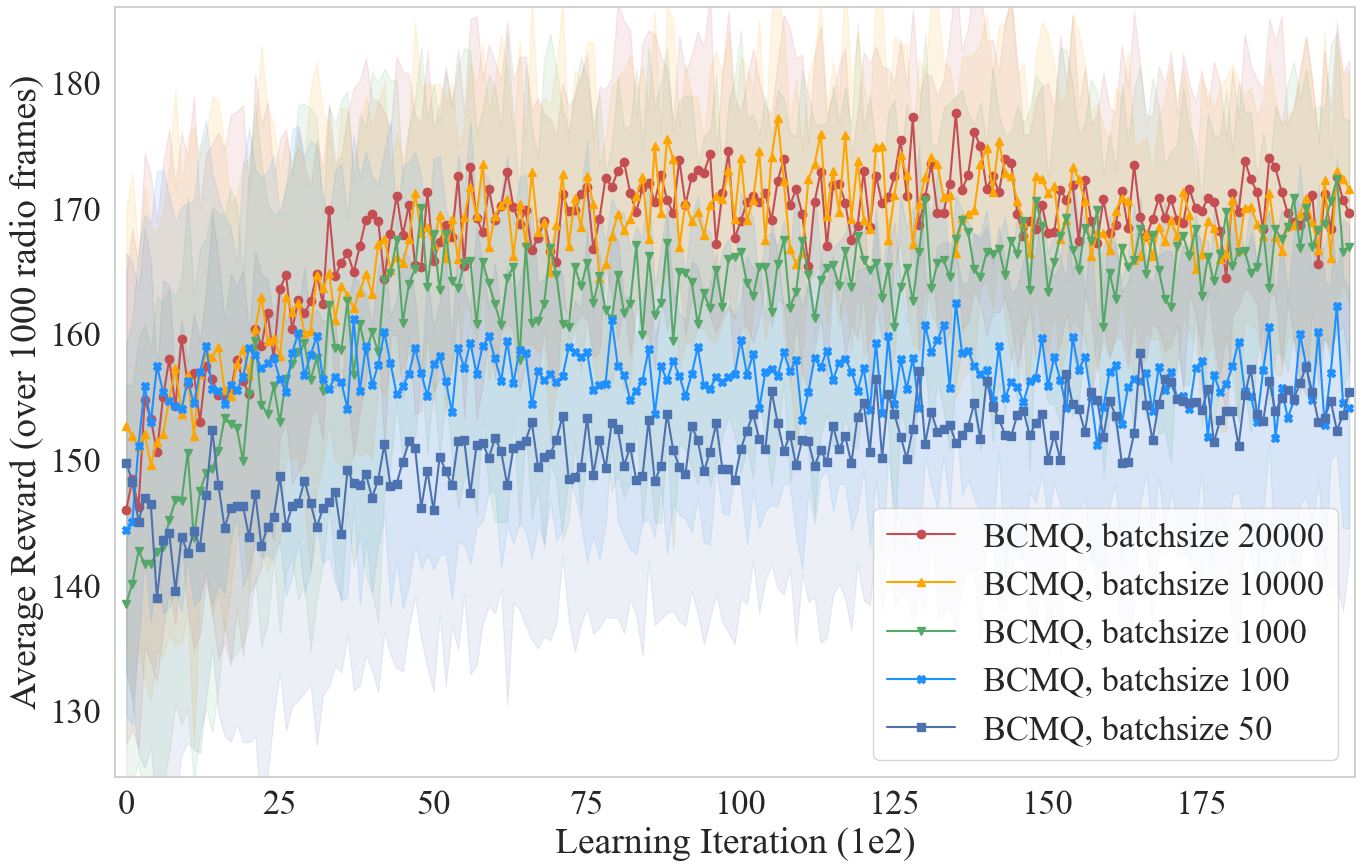}
 		\caption{Performance comparison according to batch size. High learning performance can be expected only when a sufficient number of samples is guaranteed.}
     	\label{average_reward_vs_learning_iter_CBdiffer}
    \end{figure}
    
        \textbf{Size of data.\ } In Fig. \ref{average_reward_vs_learning_iter_CBdiffer}, we plot the performance of our algorithm for different \textit{batch sizes} \footnote{This is not to be confused with the mini-batch size of the SGD algorithm used for training}. We define the batch size as the size of the given data, from which we can sample a mini-batch to train the RL agent. For our experiment, we consider the batch sizes $\{20,000, 10,000, 1000, 100, 50 \}$. We can see from the  plot that the higher the batch size, the more representative and rich the dataset is and the better the performance. But we also see that the performance improvement beyond 1000 samples is very negligible and the final convergence is very close to that of a batch size of $10,000$ or $20,000$. This supports our argument that while it might take a large number of iterations for the RL agent to converge, by taking advantage of the nature of the problem we can achieve that with only a limited amount of data. This is especially very crucial for settings like wireless networks where having access to an accurate simulator and hence large amount of data is very difficult and expensive. We also note that having a very small batch size like $50$ or $100$ is often not sufficient to learn a meaningful policy and this threshold for the batch size should be decided intelligently based on the complexity of the network under consideration. Thus, these experimental results show that batch-constrained learning approaches can be deployed for the parameter optimization of network control problems with a suitable batch size that can be adapted based on the network. 

    \begin{figure}
        \centering
 		\includegraphics[width=1.0\linewidth]{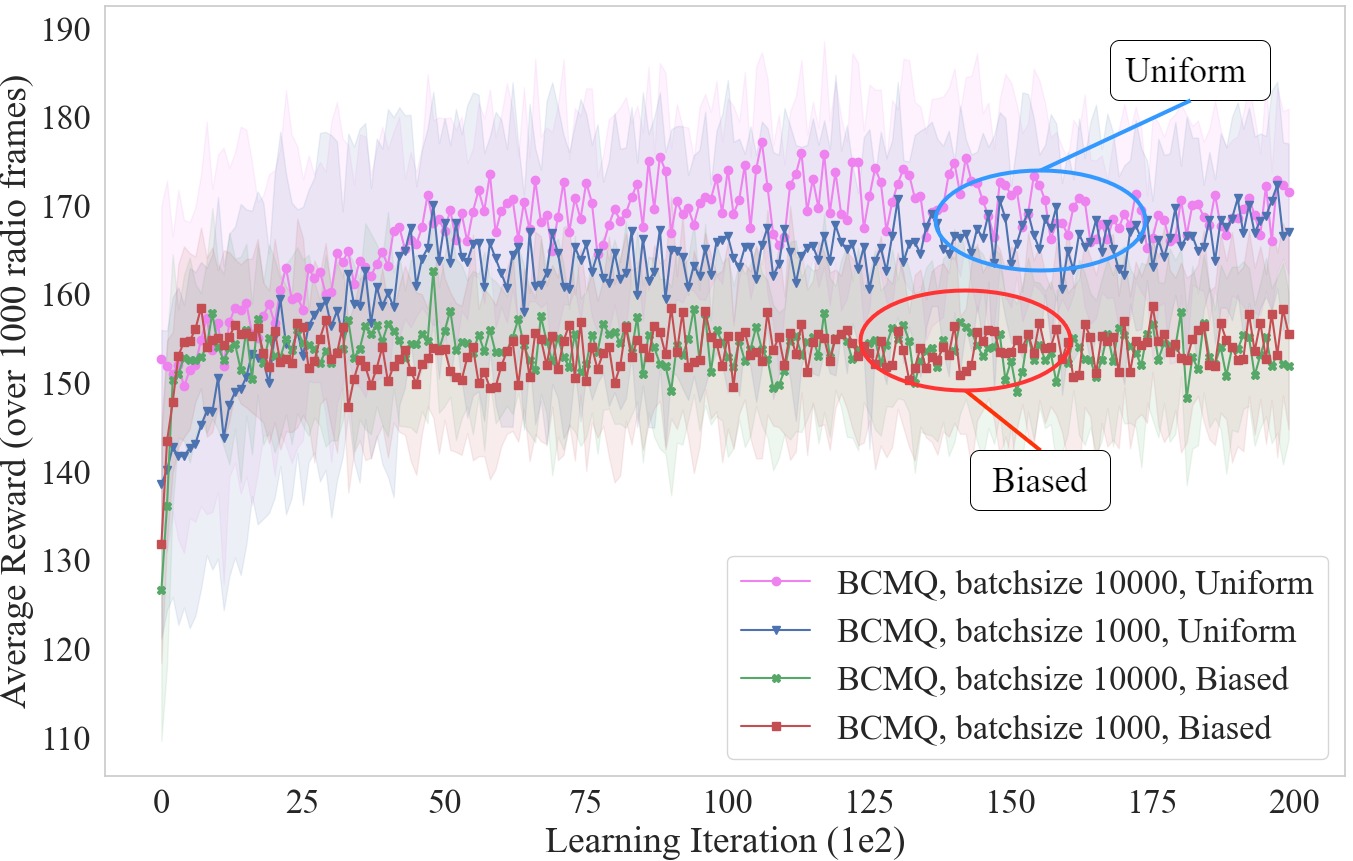}
 		\caption{Performance comparison according to data quality. To create a biased batch, the batch was configured using only 1/4 of the allowed actions. If you use a batch created with only a few actions, you cannot expect a big performance improvement.}
     	\label{fig_dataset_quality_comparison}
    \end{figure}

        \textbf{Quality of data.\ } In Fig. \ref{fig_dataset_quality_comparison}, we study the effect of the quality of data on the performance of the BCMQ algorithm. In a wireless communication system, if multiple BSs do not cooperate with each other, each BS naturally allocates a higher amount of power to a UE with low communication quality. This cooperation is one of the key metrics for deciding the usefulness of the data. To study this more, we consider two different scenarios. The first is the case where the BSs do not cooperate with each other at all and the second is the case where the BSs cooperate with each other, which is what we have considered in the previous simulations. We call the data collected from the former scenario as \textbf{biased batch} and the data from the latter scenario as \textbf{uniform batch}. We can see statistically that the uniform batch is more representative of the practical scenario and also covers more of the state-action space making it of \textit{higher quality}.

        
        It is noteworthy that the agent trained from the biased batch shows only a small performance improvement, even if the amount of data is sufficient. BCMQ can deliver the performance of the latest batch reinforcement learning by adding a 1-step rollout policy to the batch-constrained Q-Learning technique. Nevertheless, if the quality of the underlying data is seriously bad due to the biased samples, this observation is natural because the agent has no information about the transition dynamics.

    \subsection{Results Summary}
    We have demonstrated the following remarkable results through the experiments conducted in this work.
    \begin{itemize}
        \item We were able to train the reinforcement learning agent with only limited data without interaction with the environment, and surpassed the performance of the existing DQN-based method \cite{mismar2019deep}. More notably, even though the given data was generated randomly, a significant performance improvement was achieved.
        \item Given a large enough dataset, we have shown that the batch-constrained offline approach is often sufficient to achieve the same performance as that of exploration-based methods. However, we have also seen that if the size of the dataset provided is below a certain critical threshold, no amount of training can improve the performance of the RL agent.
        
        \item Performance is also significantly affected by the quality of the dataset as well as the number of training data samples for reinforcement learning.
    \end{itemize}
    Based on these observations, we conclude our project and discuss its future direction in the next section.

\section{Conclusion and Future Directions}
        Through this project, we proposed an algorithm that maximizes the SINR in radio frames of a 5G network by learning an agent without interacting with the environment. Commercial network providers can not accept the high cost of failure associated with the deployment of a sub-optimal policy, which makes the \textit{interaction less} learning very attractive.  
        
        Similar to the previous RL-based approaches, we formulated the large and computationally expensive non-convex optimization problem as a learning-based sequential problem. But we are different and better than the existing approaches in two key metrics. The first is the data sampling efficiency and the second is the lack of need for exploration. Moreover, even with these two constraints, we have shown that our proposed approach outperforms the existing DQN baseline. 
    
        In order to understand and interpret the results more, we conduct experiments to study the importance of different key factors of the problem. We demonstrate the importance of the diversity of the data and show that even with the help of the recent sophisticated SOTA algorithms such as SAC, and BCQ with rollout, we cannot achieve the best performance if the data is not good to begin. However, we show that even if only a small number of samples that cover the region of possible state domain are available, they can be used to their full potential to achieve high performance without exploration. We have experimentally shown that even a relatively small number of datasets (1000 samples) can lead to significant performance improvements. It is expected that our method can be applied for optimization in various network environments.
        
    
        Future research directions include applying our methodology in more diverse network environments. This methodology can be used not only for 5G networks but also for optimization of existing 4G networks that require interference control. Furthermore, if the environment in which more detailed information can be obtained from the UE is assumed, our approach will show higher performance. For example, if channel information can be transmitted accurately, we can choose an action that is closer to the optimum. In a more advanced communication system, network parameter optimization problems are expected to be effectively solved by the proposed approach.



\bibliographystyle{IEEEtran}
\bibliography{references.bib}
\end{document}